\documentclass{article}

\PassOptionsToPackage{numbers, compress}{natbib}

\usepackage[preprint]{neurips_2023}

\usepackage[utf8]{inputenc} 
\usepackage[T1]{fontenc}    
\usepackage{hyperref}       
\usepackage{url}            
\usepackage{booktabs}       
\usepackage{amsfonts}       
\usepackage{nicefrac}       
\usepackage{microtype}      
\usepackage{xcolor}         
\usepackage{amsmath}        
\usepackage{amsthm}
\usepackage{graphicx} 
\usepackage{subcaption} 

\title{Cooperative Backdoor Attack in Decentralized Reinforcement Learning with Theoretical Guarantee}

\author{
    Mengtong Gao\\
    Shandong University\\
    \texttt{202122300412@mail.sdu.edu.cn}\\
    \And
    Yifei Zou\\
    Shandong University\\
    \texttt{yfzou@sdu.edu.cn}\\
    \And
    Zuyuan Zhang\\
    The George Washington University\\
    \texttt{zuyuan.zhang@gwu.edu}\\
    \And
    Xiuzhen Cheng\\
    Shandong University\\
    \texttt{xzcheng@sdu.edu.cn}\\
    \And
    Dongxiao Yu\\
    Shandong University\\
    \texttt{dxyu@sdu.edu.cn}\\
}

\theoremstyle{definition}

\theoremstyle{plain}
\newtheorem{theorem}{Theorem}[section]

\theoremstyle{remark}

\begin{document}

\maketitle

\begin{abstract}
The safety of decentralized reinforcement learning (RL) is a challenging problem since malicious agents can share their poisoned policies with benign agents. The paper investigates a cooperative backdoor attack in a decentralized reinforcement learning scenario. Differing from the existing methods that hide a whole backdoor attack behind their shared policies, our method decomposes the backdoor behavior into multiple components according to the state space of RL. Each malicious agent hides one component in its policy and shares its policy with the benign agents.  When a benign agent learns all the poisoned policies, the backdoor attack is assembled in its policy. The theoretical proof is given to show that our cooperative method can successfully inject the backdoor into the RL policies of benign agents. Compared with the existing backdoor attacks, our cooperative method is more covert since the policy from each attacker only contains a component of the backdoor attack and is harder to detect. Extensive simulations are conducted based on Atari environments to demonstrate the efficiency and covertness of our method. To the best of our knowledge, this is the first paper presenting a provable cooperative backdoor attack in decentralized reinforcement learning. 
\end{abstract}

\section{Introduction}

As an important branch in safe reinforcement learning (RL), the backdoor policy attack and defense has become an important research topic~\cite{nguyen2021deep, uprety2020reinforcement, DBLP:conf/nips/KimSPS23, DBLP:conf/nips/JiaYSNRRLP23, DBLP:conf/nips/OzisikT20, DBLP:conf/icml/LinTWYMXWW23, DBLP:conf/icml/WachiS20}. Specifically, the backdoor policy in RL refers to a policy that behaves like an optimal one under a normal environment but performs poorly or acts in a specific manner when the trigger is activated in an adversarial environment. Some relevant works include the backdoor attack mechanisms~\cite{Kiourti2020TrojDRLEO, Wang_2021, gong2022baffle, cui2023badrl, liu2017trojaning, zhao2020blackbox} in the scenarios of maze environments, image recognition and autonomous driving, and the defense strategies~\cite{bharti2022provable, nguyen2021flame, shen2016auror}. Whereas, most of the works mentioned above are considered for a single RL agent and founded on numerical experiments~\cite{liu2017trojaning, zhao2020blackbox}. 
To the best of our knowledge, few of the existing works consider the backdoor attack in decentralized reinforcement learning scenarios.

This paper investigates the backdoor attack problem in decentralized reinforcement learning. First of all, our investigation is significant because Decentralized RL has broad applications, in reality, ~\cite{leottau2018decentralized}. With multi-agents to cooperatively explore an unknown environment, decentralized RL has a faster speed in finding the optimal policy than RL in a single agent. Secondly, our study is necessary because the decentralized nature makes it hard 
to verify the trustworthiness of participating agents, which opens a
door to the backdoor policy attacks. 

\begin{figure}[htbp]
    \centering
    \includegraphics[width=\textwidth]{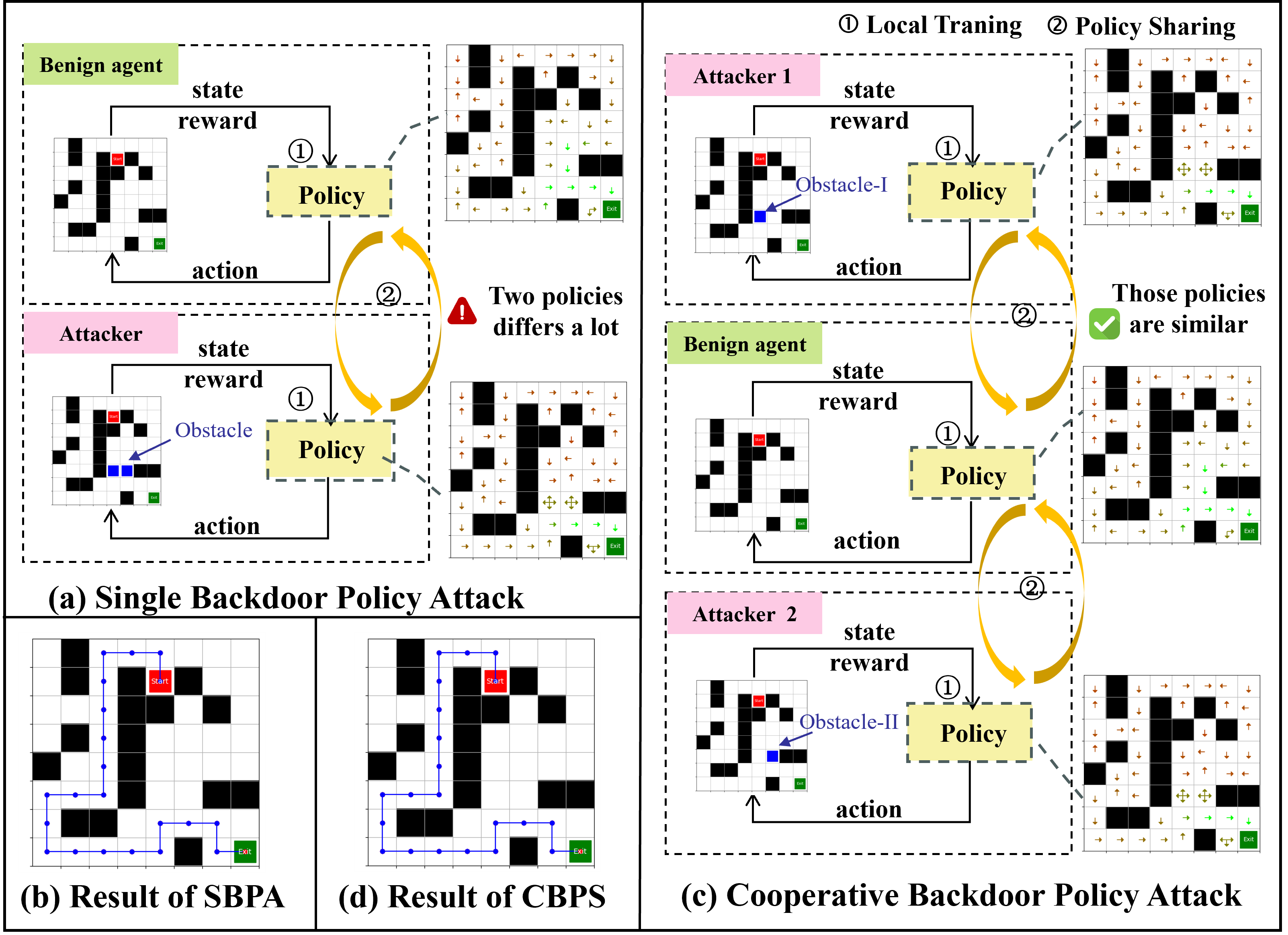}
    \caption{We study cooperative backdoor policy attacks in decentralized RL. Differing from the single backdoor policy attack that hides a whole backdoor knowledge behind its malign policy, our method decomposes the backdoor behavior into multiple components, each of which is hidden by an individual attacker within its malign policy. When a benign agent learns all the poisoned policies, the backdoor attack is assembled in its policy. Compared with a single backdoor policy attack, our method has the same attacking performance but is harder to detect.}
    \label{fig:sub1}
\end{figure}

\textbf{Demo.} A demo based on the maze environment is conducted as the motivation for our work. In our demo, the benign agent tries to find the shortest path from the maze environment. Whereas, in a backdoor maze environment, an invisible obstacle is placed to block the shortest path. The backdoor attack is defined as follows. As the player gets close, the obstacle automatically appears to block further progress. Conversely, when the player moves away, the obstacle disappears, restoring the environment to its original unobstructed state. In Fig.~\ref{fig:sub1} (a)-(d), we present the single backdoor policy attack (SBPA) with its attacking result, and the cooperative backdoor attack (CBPA) with its attacking result, respectively.

In Figure~\ref{fig:sub1} (a), the benign agent learns its policy from the benign maze, while the attacker obtains a backdoor policy from the backdoor maze with an obstacle. By exchanging their learning policies, such a backdoor knowledge will be injected into the benign agent, i.e. the benign agent will choose a longer path even in a benign maze environment, as is illustrated in Figure~\ref{fig:sub1} (c). This is because the benign agent learns the existence of the invisible obstacles (wrong knowledge) from the attacker, even though such an obstacle does not exist in the current benign environment. One concern for such an SBPA  is that the backdoor policy differs a lot from the benign policy due to the existence of an invisible obstacle. Thus, the backdoor policy is likely to be detected and denied before it is learned by a benign agent. 

In Fig~\ref{fig:sub1} (b), we propose a more covert cooperative backdoor policy attack (CBPA). Specifically, the obstacle in Fig~\ref{fig:sub1} (a)
is divided into two parts, the left and right of which are named obstacle-I and obstacle-II for short. The attacker 1 has the obstacle-I on its shortest path and the attacker 2 has the obstacle-II on its shortest path. Since the short path on the backdoor maze environment is not fully blocked, the backdoor policies are similar to the policy learned by the benign agents themselves, which makes the backdoor policies more likely to be accepted by the benign agents. When the benign agents learn the backdoor policies from attackers 1 and 2, full backdoor knowledge is assembled in its policy. 

According to the results in Fig.~\ref{fig:sub1} (c) and (d), CBPA has the same attacking results as the SBPA, i.e., the benign agent no longer chooses the shortest path even though there is no obstacle in the maze. The advantage of CBPA is that its backdoor policies are harder to detect than those in SBPA. This demo shows the feasibility of injecting some malicious knowledge into a benign agent in decentralized RL via the cooperative attack and its advantage of covertness, which motivates our work. A more general and formal definition of the backdoor attack in decentralized RL is presented in Sec.~\ref{sec.31}.

\textbf{Contribution.} In this paper, we investigate the backdoor attack problem (also known as the Trojan attack problem) in decentralized reinforcement learning. Considering that an RL policy with a full backdoor hidden behind takes the risk of being detected, a cooperative backdoor attack method, named \textit{Co-Trojan}, is proposed in our work. In our method, malicious agents decompose a given backdoor into multiple fragments according to the state space of RL. Then, each malicious agent trains its RL policy that hides a fragment behind. Compared with the RL policy containing the whole backdoor, the policy only with a backdoor fragment is less destructive and therefore unlikely to be detected. When all the policies containing the backdoor fragments are learned by the benign agents, the full backdoor will be assembled and injected into the RL policy of benign agents. The theoretical proof is given to show that  
our approach can successfully inject a whole backdoor into the RL policy of benign agents in a cooperative manner. Numerical experiments are also conducted to validate our theoretical results.

\section{Related work}
In recent years, the backdoor attack problem has become a hot research topic, with several studies addressing both the implementation and defense of such attacks within reinforcement learning~\cite{Kiourti2020TrojDRLEO,Wang_2021,gong2022baffle,cui2023badrl}, federated learning~\cite{nguyen2021flame, bagdasaryan2020backdoor, li2020federated, DBLP:conf/nips/NguyenNTDW23, DBLP:conf/nips/ZhangJCLW23, DBLP:conf/aaai/LyuHWLWL023, DBLP:conf/icml/Xie0CL21, DBLP:conf/aaai/LiuZFYXM024} and decentralized learning~\cite{shen2016auror, wang2020attack, xie2019dba}. To the best of our knowledge, we seldom consider the backdoor problem under decentralized reinforcement Learning.

\textbf{Backdoor Attack and Defense in Reinforcement Learning. }
Recent advancements in backdoor attacks within reinforcement learning focus on contaminating policies to induce anomalous agent behaviors. TrojDRL~\cite{Kiourti2020TrojDRLEO} investigates the uniform injection of triggers during training in non-targeted threat models. In competitive RL contexts, \cite{Wang_2021} focuses on the strategic selection of compromised states for executing backdoor attacks. \cite{gong2022baffle} explores the efficacy of backdoor attacks by manipulating environmental dynamics to activate backdoors in critical states. BadRL~\cite{cui2023badrl} introduces a highly sparse backdoor poisoning approach, dynamically generating distinct triggers tailored to targeted observations. \cite{li2020federated} introduces an RL-based backdoor attack framework for federated learning, emphasizing the need for advanced defenses. \cite{Wang_2021} demonstrates that backdoor attacks can be activated in competitive RL systems without direct observation manipulation, expanding the scope of potential security threats. \cite{bharti2022provable} proposes defending against backdoor policies in RL by projecting observed states into a 'safe subspace' to approximate optimality. The remaining related work are~\cite{zhang2024collaborative,zou2024distributed,zhang2024modeling,zhou2022pac} .Collectively, these studies underscore the diverse methodologies and complexities of backdoor attacks and defenses, enriching our understanding and formulation of defensive strategies.

\textbf{Backdoor Attack and Defense in Decentralized/Federated Learning. }
In the field of decentralized/federated learning, several key studies advance our understanding of backdoor attacks. \cite{nguyen2021flame} introduces FLAME, a framework for injecting backdoors by modifying local model updates. \cite{bagdasaryan2020backdoor} demonstrates the severe impact of minimal malicious client data on global models. \cite{li2020federated} proposes robust aggregation methods to mitigate malicious updates. \cite{wang2020attack} presents a covert attack exploiting privacy-preserving model updates. \cite{xie2019dba} details the DBA framework for effective and stealthy backdoor injection. \cite{li2023learning} proposes a novel approach to learning and implementing backdoor attacks in federated learning systems, demonstrating significant vulnerabilities in these models. The remaining related work are~\cite{zhou2023value,mei2023mac} . These studies emphasize significant security risks, offering crucial insights for future research.

\section{Methodology}

In this section, the system model of our decentralized reinforcement learning, the problem definition of the backdoor attack problem, the description of our cooperative backdoor attack method, and the corresponding theoretical analysis are given one by one.

\subsection{System Model of Policy-based Decentralized Reinforcement Learning}
\label{sec.31}

Similar with the classic work~\cite{leottau2018decentralized}, we consider a decentralized RL system consisting of \( n \) agents \(\{ F_i \}_{i=1}^n\). The system operates over \( T \) consecutive discrete rounds, each consisting of two key steps: local training and knowledge sharing among the agents.

In the local training step, each agent \( F_i \) independently executes its policy \(\pi_i^t\), collects data through interactions with its respective Markov process \(\{ M_i \}_{i=1}^N\), updates its policy based on its observations, and optimizes its actions to maximize the desired cumulative reward \( V_{M_i}^{\pi_i} \). The current model policy \(\pi_i^t\) is updated to \(\pi_i^{t+1}\), with the optimization objective for local training defined as:

\begin{equation}
    \pi^*_i =\arg \max_{\pi_i} V_{M_i}^{\pi_i} = \arg \max_{\pi_i}  \mathbb{E}_{A \sim \pi_i \left( \cdot | s_t \right)}  \left[ \sum_{t=0}^{\infty} \gamma^t R_i(s_t, \pi_i(s_t)) \right]
\end{equation}

In the knowledge-sharing step, agents exchange information to share key experiences learned during local training. This shared information includes but is not limited to, action rewards, state transfer information, and policy updates. To protect privacy, each agent aggregates its policies, resulting in a global policy \(\pi\), and shares it with others: 

\begin{align}
    \pi & = \text{Aggregate}(\{\pi_i\}_{i=1}^N) \quad \text{with} \quad \pi(a|s) = \frac{1}{N} \sum_{i=1}^N \pi_i(a|s), \forall s\in \mathbb{S}, \forall a\in \mathbb{A}
\end{align}
To maximize the value of agents, each agent sharing its value is an intuitive solution. Whereas, it has some concerns about the privacy leakage problem. Thus, our model considers the policy-based sharing method. In Appendix A.1, we show that the policy-based sharing method is equivalent to the value-based sharing in maximizing the total value. Meanwhile, the policy-based sharing method has a better performance on privacy protection. 

The decentralized RL system aims to maximize the cumulative reward under the global policy, as its overall optimization goal.

\begin{align}
    \pi^* & = \arg \max_{\pi_i} V_{M}^{\pi_i} = \arg \max_{\pi}  \mathbb{E}_{A \sim \pi \left( \cdot | s_t \right)}  \left[ \sum_{t=0}^{\infty} \gamma^t R(s_t, \pi(s_t)) \right]
\end{align}

By repeating the local training and knowledge sharing for sufficient times, the individual local policies \(\{\pi_i\}_{i=1}^N\) and their aggregated result \(\pi = \text{Aggregate}(\{ \pi_i \}_{i=1}^N)\) converge towards a globally optimal policy \(\pi^*\).

\subsection{Problem Definition of Backdoor Attack in Reinforcement Learning}
Backdoor attacks in Reinforcement Learning are defined by a pair of tuples \((\pi^\dagger, f)\), where \(\pi^\dagger\) is the backdoor policy and \( f \) is the trigger function~\cite{bharti2022provable}. The adversary constructs a backdoor policy \(\pi^\dagger\) that behaves identically to \(\pi^*\) within the support \( T \) of \( d^{\pi^*}_M \), but differs outside of this support.

The trigger function \( f \) is adaptive and injects triggers in the \( E^\perp \) subspace of the state space \( S \), ensuring the perceived triggered states are bounded in expectation. The backdoor policy \(\pi^\dagger\) is then defined as:

\begin{equation}
    V^{\pi}_{M, f} = \mathbb{E} \left[ \sum_{t=0}^{\infty} \gamma^t R(s_t, \pi(s_t + f(s_{0:t}))) \right] = V^{\pi \circ f}_M
\end{equation}

\textbf{Safe Subspace of the States.} The adversary commences by selecting an optimal policy \( \pi^* \), which under \( M \) yields a discounted state occupancy measure, defined as \( d^{\pi^*}_M(s) = (1 - \gamma) \sum_{t=0}^{\infty} \gamma^t \Pr(s_t = s | s_0 \sim \mu, \pi^*) \). Let \( T \subset S \) denote the support of \( d_{\pi^*}^M \), where \( T \) is the smallest closed subset in \( \mathbb{R}^D \) such that \( \Pr(T) = 1 \). It is postulated that \( \mathbb{E}_{s \sim d^{\pi^*}_M}[s] = 0 \).The eigen decomposition of the state covariance matrix under \( d^{\pi^*}_M \).

\begin{equation}
    \Sigma = \mathbb{E}_{s \sim d_{M}^{ \pi * }} 
\left[ ss^T \right] =  \sum_{i=1}^D  \lambda_i u_i u_i^T ~~~
where ~ \lambda_1 \geq \cdots \lambda_d \geq \lambda_{d+1} \geq \cdots \geq \lambda_{D}
\end{equation}

We define the top $d$ eigenspace of $\Sigma$ by $E = \text{span}(\{u_i\}_{i=1}^{d})$, and its orthogonal complement by $E^\perp = \text{span}(\{u_i\}_{i=d+1}^{D})$. The projection operators for $E$ and $E^\perp$ are given by $\text{Proj}_E = \sum_{i=1}^{d} u_i u_i^\top$ and $\text{Proj}_{E^\perp} = \sum_{i=d+1}^{D} u_i u_i^\top$. We refer to $E$ as the $\text{safe subspace of the states}$.

To understand the rationale behind the definitions and how they are utilized in the context of backdoor policy attacks in decentralized reinforcement learning, let's go through the following assumptions in detail:

\textbf{Assumption 1}\label{as1} The occupancy distribution $d^{\pi^*}_M$ has a bounded support along the smallest $D - d$ eigen-subspace $E_{\perp}$, i.e., $\exists C_0 \in \mathbb{R}$ s.t. under $d^{\pi^*}_M$, $P(\{s \in S : \| \text{Proj}_{E_{\perp}}(s) \|_2 \leq C_0\}) = 1$. Further, we assume that $d^{\pi^*}_M$ is $K$-subGaussian.

The safe subspace of states $E$ is bounded, preserving the performance of the optimal policy within the bounded support $T$, and allows for the insertion of backdoors in states outside the bounded support $T$.

\textbf{Assumption 2}\label{as2} The trigger function $f : (S)^* \rightarrow E^{\perp}$ is adaptive and the adversary can only inject a trigger in the $E^{\perp}$ subspace of the state space $S$. Further, we assume that the perceived triggered states are $B$-bounded in expectation as below, $\forall \pi, \forall t \in \mathbb{N}$, $\mathbb{E}_{s_0:t \sim d^{\pi \circ f}_M} \left[ \| (s_t + f (s_{0:t})) \|_2 \right] \leq B$ where $d^{\pi \circ f}_{0:t} M$ denotes the distribution of partial state trajectory up to time step $t$ under policy $\pi$ and trigger function $f$.

We define the backdoor policy as the composition of a given policy and a trigger function, that is
\begin{equation}
    V^{\pi}_{M, f} = \mathbb{E} \left[ \sum_{t=0}^{\infty} \gamma^t R(s_t, \pi(s_t + f(s_{0:t}))) \right] = \mathbb{E} \left[ \sum_{t=0}^{\infty} \gamma^t R(s_t, \pi \circ f(s_{0:t})) \right] = V^{\pi \circ f}_M
\end{equation}

The trigger function $f$ affects the value of the agent solely through action selection. Moreover, $f$ depends on the current state and its history, combining a Markov policy $\pi$ with the trigger function $f$ results in a non-Markov policy $\pi \circ f$ ~\cite{bharti2022provable}.

\textbf{Assumption 3}\label{as3} The backdoor policy $\pi^{\dagger}$ is $L$-Lipschitz smooth, i.e., $\exists L \in \mathbb{R}$ s.t. $\forall s, s' \in S$, $\| \pi^{\dagger}(s) - \pi^{\dagger}(s') \|_1 \leq L\|s - s'\|_2$.

\subsection{Detailed Description of Co-Trojan}

In this study, we propose a cooperative backdoor attack method in decentralized RL. Unlike existing backdoor attacks that directly share a malign policy with the entire backdoor knowledge, our method decomposes the backdoor into multiple components based on the state space of the environment. Each malicious agent hides one component in its policy. When all malign policies are learned by a benign agent, the backdoor attack is assembled in its policy.

We propose subdividing a comprehensive backdoor attack target into multiple sub-attack targets. The global backdoor attack strategy \((\pi^\dagger, f)\) is decomposed into \( N \) uncorrelated subspaces \( E_i^\perp \) for each local trigger function \( f_i \). This decomposition ensures the overall backdoor effect while maintaining stealthiness by limiting the dimension of the non-secure subspace \( E_i^\perp \).

We define the global backdoor policy \(\pi^\dagger\) and the trigger function \(f\) as follows. The trigger function \( f \) is decomposed into multiple components:
\begin{equation}
    E_i^\perp = \text{span}(\{u_i\}_{i=t_{i-1}+1}^{t_i}), t_0 = d
\end{equation}
\begin{equation}
    f_i = \Phi_f(i), f_i : (S)^* \rightarrow E_i^\perp
\end{equation}

Each local trigger function \( f_i \) is designed to operate within a specific subspace \( E_i^\perp \), orthogonal to the secure subspace \( E \). This ensures that each agent's trigger function affects only its designated subspace, preserving the overall stealthiness of the attack.

The global backdoor policy \(\pi^\dagger\) and its decomposition are defined by:
\begin{equation}
    f = \frac{1}{N} \sum_{i=1}^N f_i
\end{equation}
meaning the global trigger function \(f\) as the average of the individual local trigger functions \(f_i\) from each agent. We can get the following formula by implanting it into the attack target, i.e., the policy$\pi$ of each target:
\begin{equation}
    \pi^\dagger(s_t) = \pi \circ f(s_{0:t}) = \pi(s_t + f(s_{0:t}))
\end{equation}
it shows how the global backdoor policy \(\pi^\dagger\) is formed by incorporating the trigger function \( f \) into the original policy \(\pi\), ensuring that the modified policy reacts to the trigger while preserving normal behavior in other scenarios.
\begin{equation}
    \pi_g^\dagger = \text{Aggregate}(\{\pi_i^\dagger\}_{i=1}^N) = \frac{1}{N} \sum_{i=1}^N \pi_i(s_t + f_i(s_{0:t}))
\end{equation}
where the aggregation of individual backdoor policies \(\pi_i^\dagger\) into a global backdoor policy \(\pi_g^\dagger\).

For the state space of the global environment, we define a secure subspace \( E \) and a subspace \( E^\perp \) orthogonal to \( E \), defined by the trigger function \( f \). Each agent has the same bounded security subspace \( E \) as the global environment, which limits the normal state space unaffected by backdoor attacks. At the same time, each agent has a different distributed trigger function \( f_i \) in a different non-secure subspace \( E_i^\perp \). To balance fully demonstrating the global backdoor effect and maintaining stealthiness, we limit the dimension of the non-secure subspace \( E_i^\perp \) affected by each agent’s backdoor.

By performing singular value decomposition and feature selection on the state covariance matrix, we decompose the subspace \( E^\perp \) where the backdoor attack trigger function \( f \) is located into \( N \) uncorrelated subspaces \( E_i^\perp \), corresponding to each distributed trigger function \( f_i \).

\subsection{Theoretical Analysis for the Correctness of Co-Trojan}
In our framework, the overall backdoor attack target on a distributed RL system can be represented as a specified global backdoor attack \((\pi^{\dagger}, f)\). In order to realize the collaborative attack, we design a trigger function decomposition strategy, which decomposes the global trigger function \(f\) into \(m\) local trigger functions \(\{f_i\}_{i=1}^m\), with the aim of designing \(m\) different attackers to carry out the attack on all the agents \(\{F_i\}_{i=1}^n\).

Below, we argue that for any predefined global backdoor attack policy \(\pi^\dagger\), a decomposition \(\Phi_f(i)\) can be found, which, through the aggregation process of decentralized RL, can make the resulting global backdoor policy \(\pi_g^\dagger\) accurately approximate our target global backdoor policy.

Consider the global trigger function \(f\), which imposes constraints by confining the activated triggers to a subspace \(E^\perp\) formed by a series of vectors \(\{ u_i \}_{i=d+1}^D\). For each local trigger function \(f_i\), they are part of the global trigger function \(f\), and their corresponding subspace and decomposition method can be expressed as follows:
\begin{align}
    E_i^\perp & = \text{span}(\{u_i \}_{i=t_{i-1}+1}^{t_i}), t_0 = d \\
    f_i & = \Phi_f(i) , f_i : (S)^* \rightarrow E_i
\end{align}

For each local trigger's corresponding subspace can be defined as \(E_i^\perp = \text{span}(\{ u_i \}_{i=d_i}^{t_i})\), where the index \(d_i\) starts from \(t_{i-1}+1\). with \(t_0 = d\) as the initial value. In this way, we are able to realize an accurate and non-missing division of the subspace where the global trigger is located. In order to characterize the implementation of this partition, we introduce the function \(\Phi_f\), and in the following, we will show a partitioning strategy that is both theoretically defined and close to practical applications.
\begin{equation}
    f = \frac{1}{N} \sum_{i=1}^N f_i
\end{equation}

For a given backdoor attack target $(\pi^\dagger, f)$ our targeting policy is
\begin{equation}
    \pi^\dagger(s_t) = \pi \circ f (s_{0:t}) = \pi(s_t + f(s_{0:t})), 
\end{equation}
and the global backdoor policy that we have trained by aggregation $\pi_g^\dagger$ is
\begin{equation}
    \pi_g^\dagger = \text{Aggregate}(\{\pi_i^\dagger\}_{i=1}^N) = \frac{1}{N} \sum_{i=1}^N \pi_i^\dagger = \frac{1}{N} \sum_{i=1}^N \pi_i(s_t+f_i(s_{0:t}))
\end{equation}

According to the definition of decentralized RL, we can learn that the locally optimal policy $\pi_i^*$ converges to the globally optimal policy $\pi^*$. We can assume that after t rounds of training, the aggregated policy $Aggregate(\{\pi_i\}_{i=1}^N)$ converges to the global optimum $\pi^*$, the

\begin{equation}
    \frac{1}{N} \sum_{i=1}^N \pi_i^{t}(s_t+f_i(s_{0:t})) \xrightarrow{t \rightarrow \infty} \frac{1}{N} \sum_{i=1}^N \pi(s_t+f_i(s_{0:t}))
\end{equation}

\begin{theorem}
    For any predefined global backdoor attack policy \(\pi^\dagger\), a decomposition \(\Phi_f(i)\) can be found, which, through the policy sharing process of decentralized RL, can make the resulting global backdoor policy \(\pi_g^\dagger\) accurately approximate our target global backdoor policy.
\end{theorem}
Based on the above theorem, we conclude that backdoor attacks based on decentralized RL can have guaranteed performance even with the use of distributed computing acceleration.

\begin{proof}
To demonstrate that the distributed training results are close to the target, we must establish the following inequality:
\begin{equation}
\left\| \pi(s_t + f(s_{0:t})) - \frac{1}{N} \sum_{i=1}^N \pi(s_t + f_i(s_{0:t})) \right\| \leq \text{const}
\label{18}
\end{equation}
First, using the L-Lipschitz condition~\ref{as3} for $\pi$ and the triangle inequality for norms, we have
\begin{align}
    \left\| \pi(s_t + f(s_{0:t})) - \frac{1}{N} \sum_{i=1}^N \pi(s_t + f_i(s_{0:t})) \right\| 
    &= \frac{1}{N} \left| \left| N \pi(s_t + f(s_{0:t})) - \sum_{i=1}^N \pi(s_t + f_i(s_{0:t}))  \right| \right| \ \\
    &\leq \frac{1}{N} \sum_{i=1}^N L \left\| f(s_{0:t}) - f_i(s_{0:t}) \right\|
\end{align}
Next, we consider the relationship between $f$ and $f_i$. Given that $f$ is the average of $f_i$, for any $f_i$, we have
\begin{align}
    \left\| f(s_{0:t}) - f_i(s_{0:t}) \right\| 
    &\leq \frac{1}{N} \sum_{j=1}^N \left\| f_j(s_{0:t}) - f_i(s_{0:t}) \right\|
\end{align}
To further simplify this inequality, we assume that the norms of $f$ and $f_i$ are bounded. Specifically, we assume that the expected squared norm is bounded~\ref{as2}
\begin{equation}
    E\left[ \|s_t + f(s_{0:t})\|^2 \right] \leq B
\end{equation}
And $f_i(s_{0:t})$ is an instance in the specified state $s_t$, which still satisfies the bounded condition
\begin{equation}
    \left\| f_j(s_{0:t}) - f_i(s_{0:t}) \right\| \leq 2B
\end{equation}
So, for all \(i\) , we have
\begin{equation}
    \left\| f(s_{0:t}) - f_i(s_{0:t}) \right\| \leq 2B
\end{equation}
Finally, combining these results, we conclude
\begin{equation}
    \left\| \pi(s_t + f(s_{0:t})) - \frac{1}{N} \sum_{i=1}^N \pi(s_t + f_i(s_{0:t})) \right\| \leq 2LB
    \label{24}
\end{equation}

Through Inequality~\ref{24}, we can prove Inequality~\ref{18}, demonstrating that the outcome of the distributed backdoor attack is close to the overall attack target, further validating the accuracy of our method.

The detailed proof is provided in Appendix A.2.
\end{proof}

\section{Numerical Results}
In this section, we present experimental results to evaluate the effectiveness of our cooperative backdoor attack algorithm in decentralized reinforcement learning. We selected two Atari video games, Breakout and Seaquest, for our study. We train individual agents within a decentralized  RL framework using an open-source implementation of the parallel advantage actor-critic (PAAC)~\cite{Kiourti2020TrojDRLEO}. Experiments are performed on a server with an i7-13700K CPU and an NVIDIA GTX 2080Ti GPU.

\subsection{Experimental Setup}
\textbf{Breakout Environment.}
In the Breakout game, the objective is to destroy all bricks using a ball and paddle without letting the ball pass the paddle. We insert specific sequences of actions that trigger backdoors, causing the paddle to miss the ball at crucial moments. The backdoor policy is trained in a decentralized agent system, where triggers are decomposed and assigned to multiple malicious agents.

\textbf{Seaquest Environment.}
In Seaquest, players control a submarine to rescue divers and avoid enemy submarines and sea creatures. We add backdoor triggers that cause the submarine to dive uncontrollably when certain conditions are met. Similar to Breakout, the backdoor policy is trained in a decentralized manner with distributed triggers.

For both environments, the training process included:
\begin{itemize}
    \item \textbf{Local Training:} Each agent, including both benign and malicious agents, updates its local policy based on the experiences gathered in the environment.
    \item \textbf{Policy Sharing:} All agents share their policies. The malicious agents share policies containing fragments of the backdoor trigger, while benign agents share clean policies.
    \item \textbf{Policy Aggregation:} The shared policies are aggregated by each agent as a new one for further training.
    \item \textbf{Inference:} During inference, the comprehensive backdoor policy demonstrates its impact by triggering the backdoor conditions in both the breakout and the Seaquest environments.
\end{itemize}

\subsection{Numerical Results}

\begin{figure}[tbp]
    \centering
    \begin{subfigure}[b]{0.32\textwidth}
        \centering
        \includegraphics[width=\textwidth]{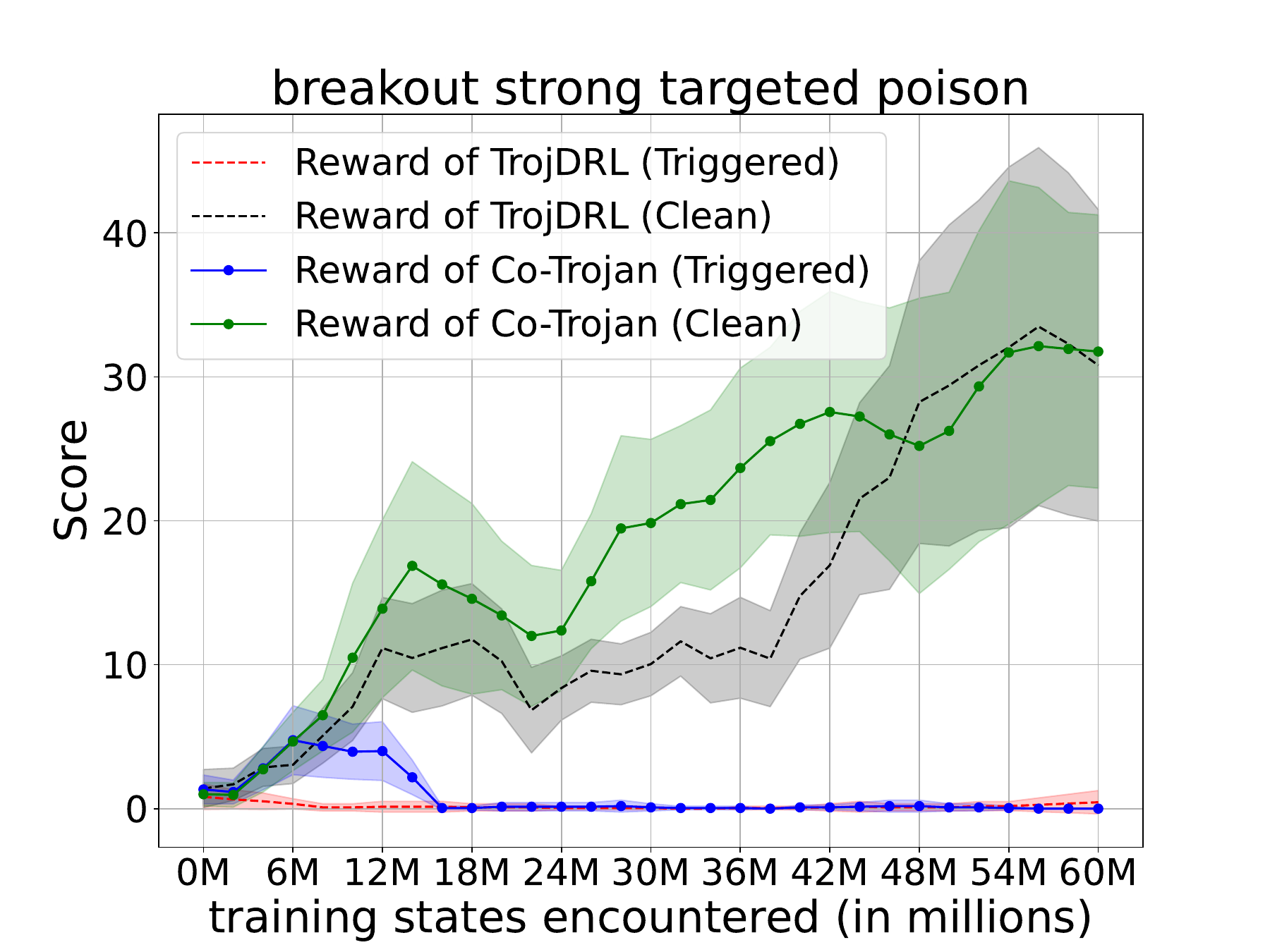}
    \end{subfigure}
    \begin{subfigure}[b]{0.32\textwidth}
        \centering
        \includegraphics[width=\textwidth]{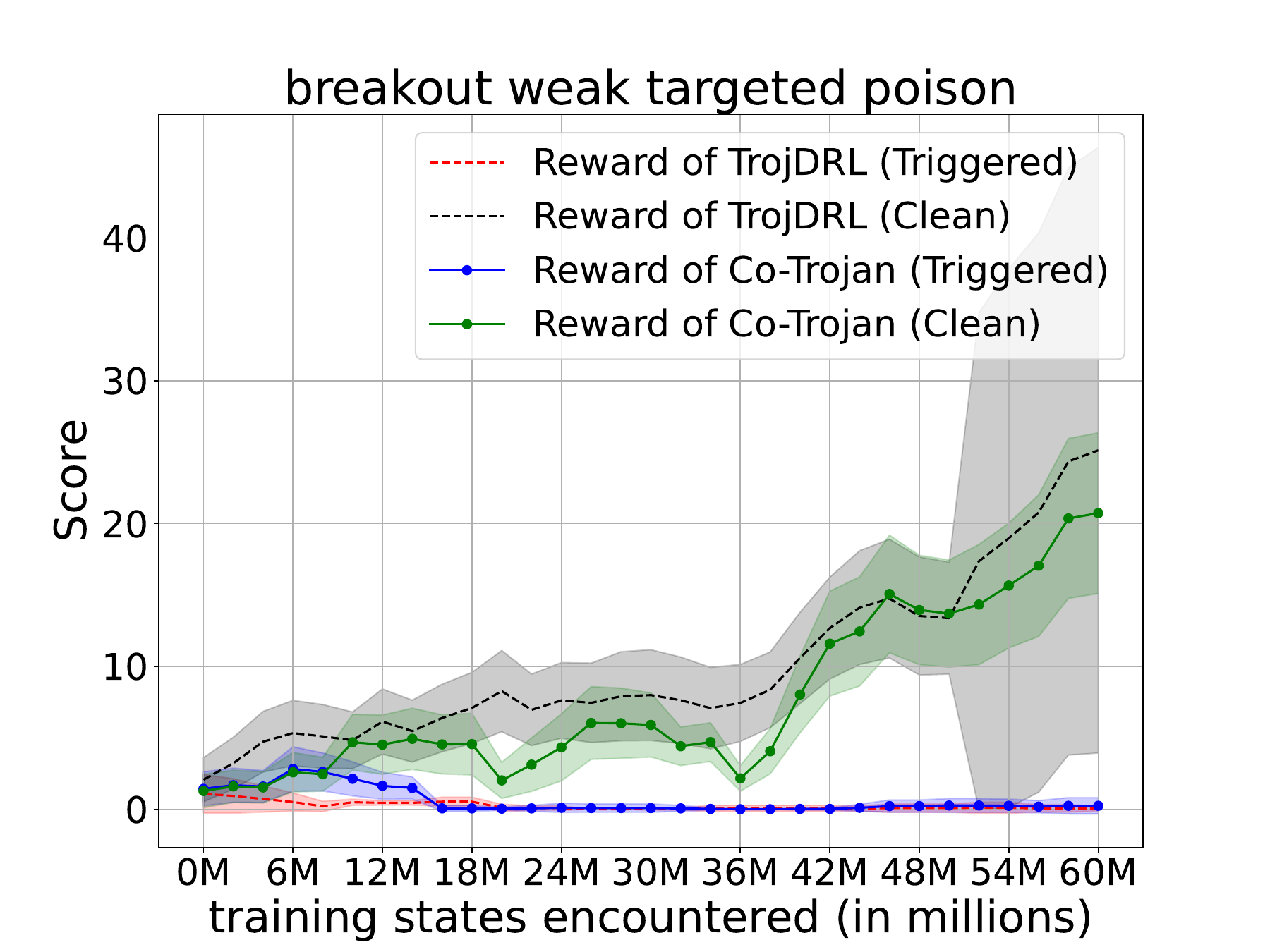}
    \end{subfigure}
    \begin{subfigure}[b]{0.32\textwidth}
        \centering
        \includegraphics[width=\textwidth]{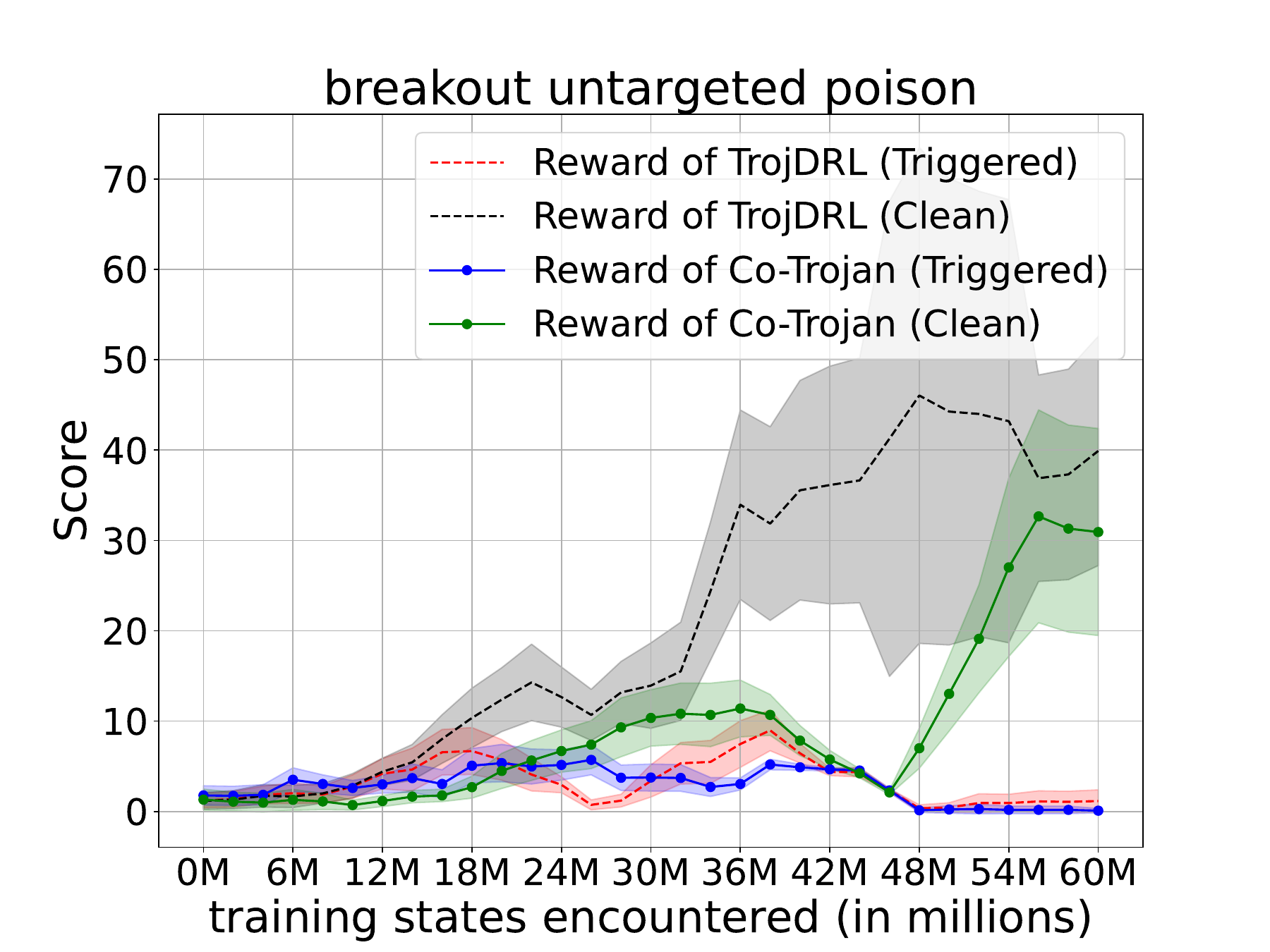}
    \end{subfigure}
    \caption{Performance Results for Breakout with Various Poisoning Conditions: (a) Strong Targeted Poison, (b) Weak Targeted Poison, and (c) Untargeted Poison. Each subplot shows the average rewards for TrojDRL (triggered), TrojDRL (clean), Co-Trojan (triggered), and Co-Trojan (clean). The lines are smoothed by averaging every five data points.}
    \label{fig:breakout_results}
\end{figure}

\textbf{Breakout Result.} Agents with the backdoor policy show a significant increase in missed balls at critical game moments, validating the effectiveness of the embedded backdoor. The performance of the backdoor policies under various poisoning conditions—strong targeted poison, weak targeted poison, and untargeted poison—is illustrated in Figure \ref{fig:breakout_results}. Specifically, the x and y-axes in Figure \ref{fig:breakout_results} represent the number of training steps and the score of the agent in the breakout game, respectively. Each subplot shows the average rewards for TrojDRL (triggered), TrojDRL (clean), Co-Trojan (triggered), and Co-Trojan (clean). The TrojDRL (triggered) and TrojDRL (clean) indicate the performances of TrojDRL in the trigger environment and the clean (non-trigger) environment, respectively. Similar denotations are given for Co-Trojan (triggered), and Co-Trojan (clean). The lines in these plots have been smoothed by averaging every five data points. 

By analyzing the Co-Trojan (clean) and Co-Trojan (triggered) performance within the same subplot, it is evident that the scores are significantly lower in the triggered environment. This observation confirms the effectiveness of our collaborative backdoor policy attack. Furthermore, when comparing the performance of TrojDRL and Co-Trojan in both triggered and clean environments within the same subplot, we find that the impact of our attack is comparable to the standard attack effect, thereby validating the accuracy of the collaborative backdoor attack. Finally, the comparison of curves across multiple subplots substantiates that our collaborative backdoor attack is effective under various poisoning conditions.

\begin{figure}[tbp]
    \centering
    \begin{subfigure}[b]{0.32\textwidth}
        \centering
        \includegraphics[width=\textwidth]{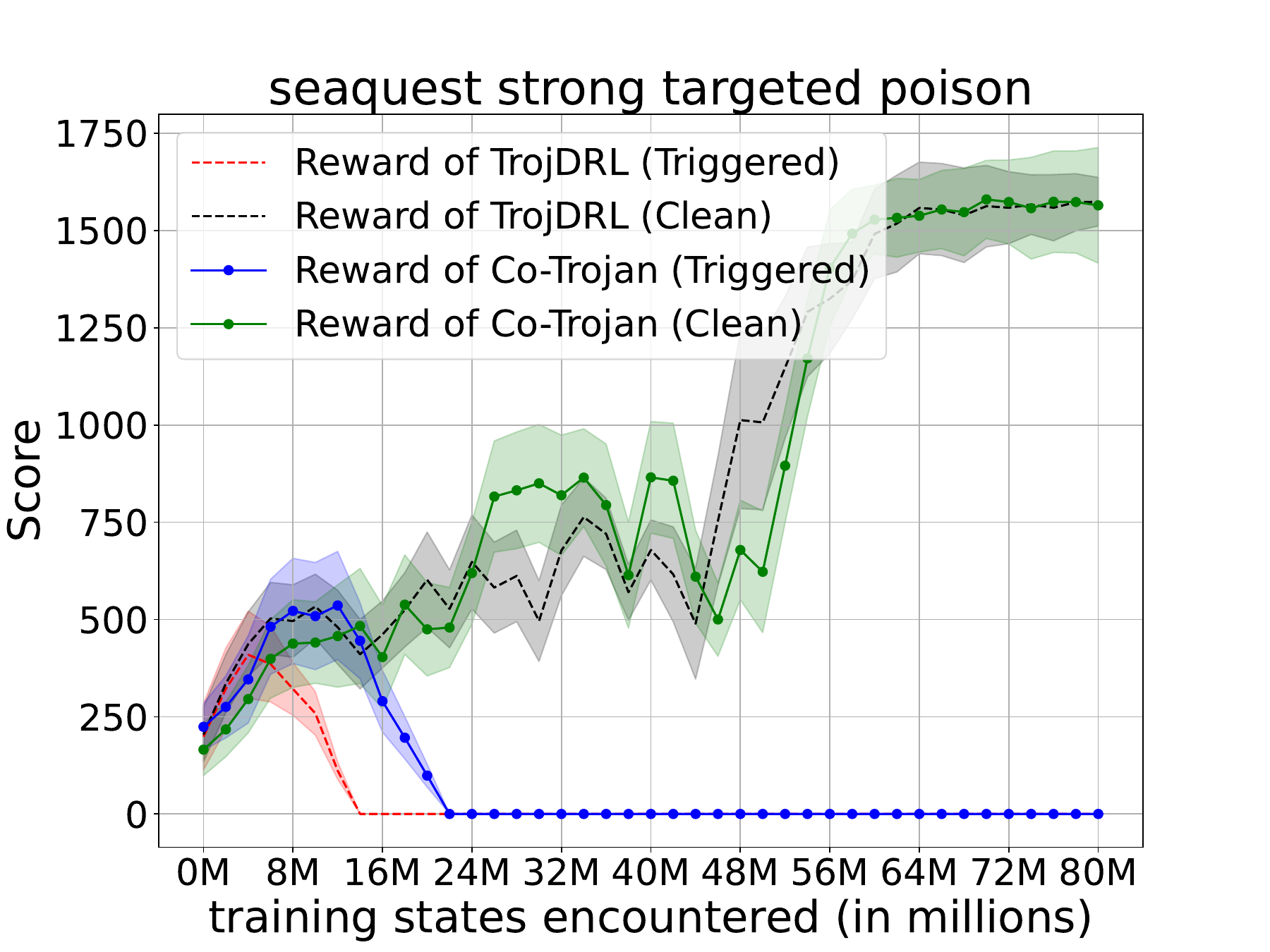}
    \end{subfigure}
    \begin{subfigure}[b]{0.32\textwidth}
        \centering
        \includegraphics[width=\textwidth]{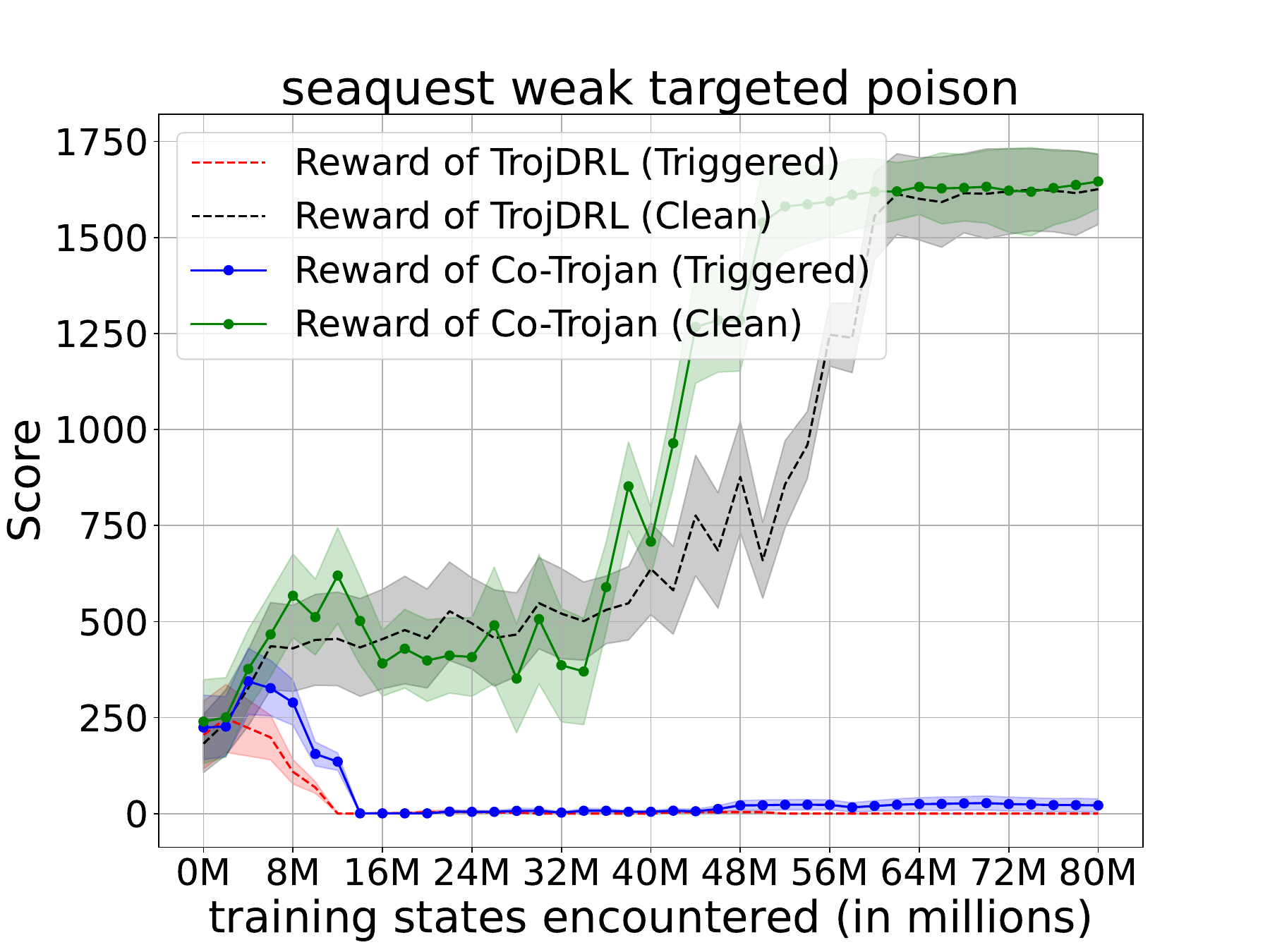}
    \end{subfigure}
    \begin{subfigure}[b]{0.32\textwidth}
        \centering
        \includegraphics[width=\textwidth]{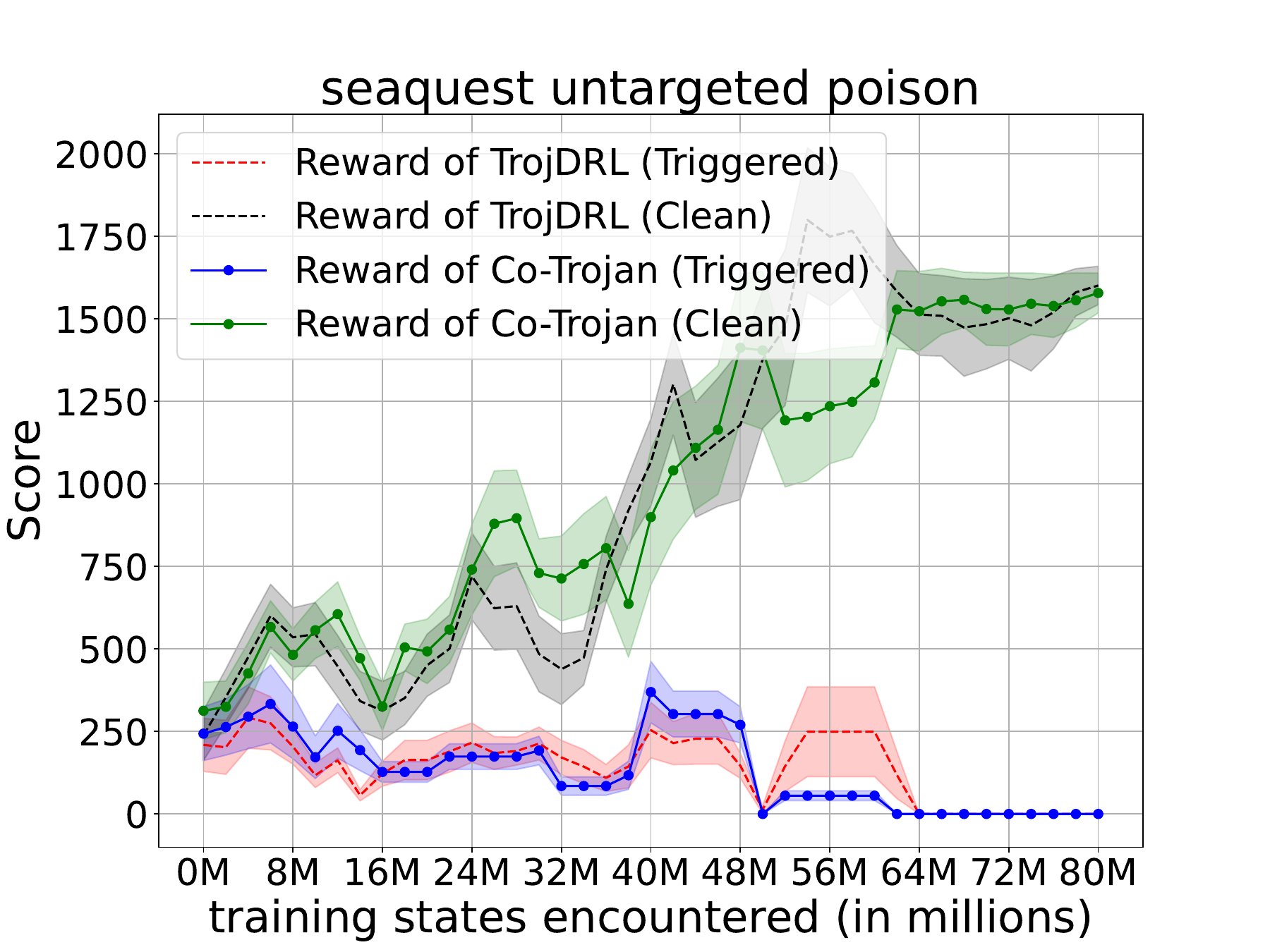}
    \end{subfigure}
    \caption{Performance Results for Seaquest with Various Poisoning Conditions: (a) Strong Targeted Poison, (b) Weak Targeted Poison, and (c) Untargeted Poison. Each subplot shows the average rewards for TrojDRL (triggered), TrojDRL (clean), Co-Trojan (triggered), and Co-Trojan (clean). The lines are smoothed by averaging every five data points.}
    \label{fig:seaquest_results}
\end{figure}

\textbf{Seaquest Result.} The submarine frequently dived uncontrollably during key game stages, demonstrating the backdoor's impact on the agent's behavior. The performance results for Seaquest under various poisoning conditions are shown in Figure \ref{fig:seaquest_results}. Similar to Breakout, each subplot shows the average rewards for TrojDRL (triggered), TrojDRL (clean), Co-Trojan (triggered), and Co-Trojan (clean). The lines are smoothed by averaging every five data points.

In the same subplot, comparing Co-Trojan (clean) and Co-Trojan (triggered) shows a significantly lower score in the triggered environment, confirming the effectiveness of our collaborative backdoor policy attack. Additionally, comparing the performance of trojDRL and Co-Trojan in both triggered and clean environments demonstrates that our attack effect is similar to the standard attack effect, confirming the correctness of the collaborative backdoor attack. By comparing the curves in multiple subplots, we verify that our collaborative backdoor attack is effective under different poison conditions.

\textbf{Summary.} The results demonstrate that our method effectively creates backdoor policies in both centralized and decentralized environments. By decomposing the backdoor trigger into smaller components and assigning them among multiple agents, we achieve effects equivalent to centralized backdoor attacks, thereby confirming the correctness of our approach in decomposing and recomposing backdoor policies to achieve the desired impact. Our experiments further validate that backdoor attacks can be effectively introduced into decentralized reinforcement learning environments, significantly enhancing the attack's covert nature and impact. Future work will focus on developing robust defense mechanisms against such decentralized backdoor attacks.

\section{Conclusion}
This paper investigates the backdoor attack problem in decentralized reinforcement learning and proposes a cooperative backdoor attack method, named Co-Trojan.
Specifically, Co-Trojan leverages the state space of the environment to decompose the backdoor into multiple components, each hidden by a malicious agent. When aggregated by benign agents, the complete backdoor attack is assembled covertly. This cooperative strategy reduces the detection risk and enhances the attack's concealment. In summary, our paper has demonstrated the existence of a provable decomposition mechanism for cooperative backdoor policy attacks based on reinforcement learning theory. Our approach provides an efficient and stealthy method for implementing backdoor attacks in decentralized reinforcement learning systems. Investigating a general backdoor defense strategy in decentralized reinforcement learning will be our work in the future.

\bibliographystyle{unsrtnat}
\bibliography{ref}






\newpage
\appendix

\section{Appendix}

\subsection{Proof of Consistency between policy aggregation and value aggregation}

\begin{proof}{Theorem 1}
In Decentralized Reinforcement Learning, each agent operates in a distinct and independent local environment, sharing only their observed experiences. This setup poses a challenge for achieving an optimal policy within the global environment \(M\) through direct training alone. Instead, it necessitates individual training in each local setting to approximate the global objective.

The rationale behind choosing policy aggregation in Decentralized RL is rooted in the need for agents to collaboratively learn and optimize their policies while maintaining privacy and efficiency. Each agent \( F_i \) trains locally, updating its policy \(\pi_i^t\) based on interactions with its specific Markov process \(\{ M_i \}_{i=1}^N\). This process continues until the local policy \(\pi_i\) approaches its optimal form, denoted as \(\pi_i^*\). The local training objective is to maximize the expected cumulative reward \( V_{M_i}^{\pi_i} \).

Following local training, agents share key experiences, including action rewards, state transition information, and policy updates. This sharing is performed through policy aggregation to protect privacy, forming a global policy \(\pi\). By aggregating policies, agents can effectively share knowledge without compromising their individual learning processes or privacy.

The overall value function \( V_{M}^{\pi} \) can be expressed as the aggregation of the locally optimized value functions:

\begin{equation}
    V_{M}^{\pi} \leftarrow \frac{1}{N} \sum_{i=1}^N V_M^{\pi_i}(s_t)
\end{equation}

This aggregation ensures that the globally optimal value function is a reflection of the locally optimized value functions, aligning with our goal of finding the optimal policy. Each agent's value function \(V\) remains private, and the policy \(\pi_i\) cannot be directly inverted to reveal private data. This aspect emphasizes the importance of policy aggregation in preserving privacy while achieving collaborative optimization.

We propose the following theorem to formalize the correctness and necessity of policy aggregation:
\begin{theorem}
In the Decentralized Reinforcement Learning system, the training effects of the aggregated policy function and the aggregated value function are equivalent.
\end{theorem}
This theorem asserts that aggregating policy functions and value functions leads to equivalent training outcomes, validating the approach of policy aggregation. This equivalence is crucial for ensuring that the decentralized framework can achieve a globally optimal policy through collaborative local training and experience sharing.

In decentralized environments, where each agent operates within a unique and isolated local environment, the sharing of experiences is restricted to observed interactions. Consequently, formulating an optimal policy within the overarching global environment, denoted as \(M\), transcends the bounds of direct training methodologies. Instead, it necessitates the individual training of agents within their respective local settings, aimed at closely approximating the global objective. This approach is premised on the notion that maximizing the value obtained by each client through localized training efforts indicates proximity to their respective optimal policies. Under such circumstances, when local policies are nearly optimal, the aggregate global policy, formulated as \(\pi\), approximates the global optimum, thereby maximizing rewards across the entire system.

This relationship can be formally expressed as:

\begin{align}
    V_{M}^{\pi} 
    & \leftarrow \frac{1}{N} \sum_{i=1}^N V_M^{\pi_i}\\
     & = \frac{1}{N} \sum_{i=1}^N V_M^{\pi_i} (s_t) \\ 
     &= \frac{1}{N} \sum_{i=1}^N \mathbb{E}_{A \sim \pi_i \left( \cdot | s_t \right)}  \left[ \sum_{t=0}^{\infty} \gamma^t R_i(s_t, \pi_i (s_t))) \right]\\
     &= \sum_{t=0}^{\infty} \gamma^t \sum_{a \in A} \frac{1}{N} \sum_{i=1}^N  \pi_i(a | s_t ) R_i(s_t, a)
\end{align}
where \(V_{M}^{\pi}\) represents the expected value under the global policy \(\pi\), and \(V_{M_i}^{\pi_i}\) denotes the expected value under the local policy \(\pi_i\) for the \(i\)-th agent at state \(s_t\). This framework underscores the importance of individual optimization in local settings as a strategy to enhance collective performance in a distributed system.

Referring to the definition of aggregation in DRL, we can obtain the global strategy trained after aggregation:
\begin{align}
    \sum_{t=0}^{\infty} \gamma^t \sum_{a \in A} \frac{1}{N} \sum_{i=1}^N  \pi_i(a | s_t) R_i(s_t, a)
    & \xrightarrow{Aggregate(\pi_i)} 
    \sum_{t=0}^{\infty} \gamma^t \sum_{a \in A} \pi_g(a | s_t) R(s_t, a) \\
    & = \mathbb{E}_{A \sim \pi_g \left( \cdot | s_t \right)}  \left[ \sum_{t=0}^{\infty} \gamma^t R(s_t, \pi_g (s_t)) \right] \\
    & = V_{M_g} ^ {\pi_g}
\end{align}
\end{proof}

We also refer to the following work~\cite{wang2023osteoporotic,fang2022coordinate,fang2023implementing,mei2022bayesian,mei2024projection,chen2023real,zhou2023double,chen2021bringing}

\subsection{Proof of Correctness}

\begin{proof}[Theorem 3.1]
    To prove that the distributed training results are close to the target, it is sufficient to prove the following inequality,
\begin{equation}
    \left| \left| \pi(s_t + f(s_{0:t})) - \frac{1}{N} \sum_{i=1}^N \pi(s_t + f_i(s_{0:t})) \right| \right| \ \leq const
\end{equation}
Based on Assumption3~\ref{as3} that the strategy $\pi$ satisfies the L-Lipschitz condition and the vector paradigm satisfies the triangular inequality, we can have the following derivation that

\begin{align}
    \left| \left| \pi(s_t + f(s_{0:t})) - \frac{1}{N} \sum_{i=1}^N \pi(s_t + f_i(s_{0:t})) \right| \right| \ 
    &= \frac{1}{N} \left| \left| N \pi(s_t + f(s_{0:t})) - \sum_{i=1}^N \pi(s_t + f_i(s_{0:t}))  \right| \right| \ \\
    &= \frac{1}{N} \left| \left| \sum_{i=1}^N \left[ \pi(s_t + f(s_{0:t})) - \pi(s_t + f_i(s_{0:t})) \right] \right| \right| \ \\
    &\leq \frac{1}{N} \sum_{i=1}^N \left| \left| \pi(s_t + f(s_{0:t})) - \pi(s_t + f_i(s_{0:t})) \right| \right| \ \\
    &\leq \frac{1}{N} \sum_{i=1}^N L \left| \left| s_t + f(s_{0:t}) - s_t + f_i(s_{0:t})) \right| \right| \ \\
    &= \frac{1}{N} \sum_{i=1}^N L \left| \left| f(s_{0:t}) - f_i(s_{0:t}) \right| \right| \ \\
\end{align}
It is known that $f = \frac{1}{N} f_i$ for $\left| \left| f(s_{0:t}) - f_i(s_{0:t}) \right| \right| \ $ with the following derivation.
\begin{align}
    \left| \left| f(s_{0:t}) - f_i(s_{0:t}) \right| \right| 
    &= \left| \left| \frac{1}{N} \sum_{j=1}^N f_j(s_{0:t}) - f_i(s_{0:t}) \right| \right| \\
    &= \left| \left| \frac{1}{N} \sum_{j=1}^N \left[ f_j(s_{0:t}) - f_i(s_{0:t})  \right] \right| \right| \\
    &\leq \frac{1}{N} \sum_{j=1}^N \left| \left| f_j(s_{0:t}) - f_i(s_{0:t})\right| \right| \
\end{align}

Below we show that $\left| \left| f(s_{0:t}) - f_i(s_{0:t}) \right| \right|$ is bounded.
Assumption2~\ref{as2} Suppose that, for any strategy \(\pi\) and any natural number \(t\), when the state \(s_0, s_1, ... , s_t\) is sampled according to the combined distribution of the strategy \(\pi\) and function \(f\) \(d_{\pi \circ f, 0:t}\) satisfies \(E\left[ \|s_t + f(s_{0:t})\|^2 \right] \leq B\). \(f(s_{0:t})\) denotes the cumulative effect of the function \(f\) from the start state to the current state, and the expectation of the squared paradigm of \(s_t + f(s_{0:t})\) is bounded. And $f_i(s_{0:t})$ is an instance in the specified state $s_t$, which still satisfies the bounded condition, i.e.

\begin{align}
   \left| \left| f_j(s_{0:t}) - f_i(s_{0:t})\right| \right| \
   &= \left| \left| s_t +f_j(s_{0:t}) - (s_t + f_i(s_{0:t}))\right| \right| \ \\
   &\leq \left| \left| s_t +f_j(s_{0:t})\right| \right| \ + \left| \left| s_t +f_i(s_{0:t})\right| \right| \ \\
   & \leq 2B
\end{align}

So, for all \( i \), there is $ \left| \left| f(s_{0:t}) - f_i(s_{0:t}) \right| \right| \leq \frac{1}{N} \sum_{j=1}^N 2B = 2B\ $.
In summary, we can conclude that
\begin{equation}
    \left| \left| \pi(s_t + f(s_{0:t})) - \frac{1}{N} \sum_{i=1}^N \pi(s_t + f_i(s_{0:t})) \right| \right| \ \leq 2LB
\end{equation}
\end{proof}

\end{document}